%% file: main.tex
\documentclass[letterpaper, 10 pt, conference]{ieeeconf}  
\IEEEoverridecommandlockouts
\usepackage{url}
\usepackage{balance}
\usepackage{amsmath}
\usepackage{amssymb,bm}
\usepackage{amsfonts}
\usepackage{enumerate}
\usepackage{tabulary}
\usepackage{multirow}
\usepackage{cite}
\usepackage{lipsum}  
\usepackage{graphicx}
\usepackage{epstopdf}
\usepackage[misc]{ifsym}
\usepackage{color}
\usepackage{xcolor}
\usepackage{xxcolor}
\usepackage{pgf}
\usepackage{pgfplots}
\usepackage{tikz}
\usepackage{float}
\usepackage{lipsum}
\usepackage{tabularx} 
\usepackage[nomessages]{fp}

\definecolor{myBlue}{rgb}{0.5294, 0.8078, 0.92156}
\newlength{\maxlen}

\usepackage{array}
\usepackage{ragged2e}
\usepackage[font=footnotesize]{caption}
\usepackage{pifont}

\usepackage{colortbl}
\usepackage{svg}
\usepackage{booktabs}
\usepackage[ruled,linesnumbered]{algorithm2e}
\SetKwRepeat{Do}{do}{while}%
\definecolor{myGray}{rgb}{0.85,0.85,0.85}
\usepackage{verbatim}
\usepackage{hyperref}
\usepackage{makecell}
\usepackage{xfrac}

\usepgfplotslibrary{groupplots,fillbetween}
\usepackage{pgfplotstable}

\usepackage{ifthen}
\usepackage{forloop}
\usepackage{listings}

\definecolor{codegreen}{rgb}{0,0.6,0}
\definecolor{codepurple}{rgb}{0.58,0,0.82}
\definecolor{backcolour}{rgb}{0.95,0.95,0.92}
\lstdefinestyle{buzz}{
    backgroundcolor=\color{black!5},   
    commentstyle=\color{codegreen},
    keywordstyle=\color{blue},
    numberstyle=\tiny\color{black!30},
    stringstyle=\color{codepurple},
    basicstyle=\footnotesize\ttfamily,
    breakatwhitespace=false,         
    breaklines=true,                 
    captionpos=b,                    
    keepspaces=true,                 
    numbers=left,                    
    numbersep=5pt,                  
    showspaces=false,                
    showstringspaces=false,
    showtabs=false,                  
    tabsize=2,
}
\DeclareMathOperator*{\argmin}{argmin}
\DeclareMathOperator*{\argmax}{argmax}

\SetAlCapSkip{0.5em}

\title{\LARGE \bf
Uncertainty-aware Gaussian Mixture Model for UWB Time Difference of Arrival Localization in Cluttered Environments
} 

\author{
Wenda Zhao,
Abhishek Goudar, 
Mingliang Tang,
Xinyuan Qiao,
and Angela P. Schoellig
\thanks{Institute for Aerospace Studies, University of Toronto, the University of Toronto Robotics Institute, Vector Institute for Artificial Intelligence, Toronto, Canada; Technical University of Munich, Munich Institute for Robotics and Machine Intelligence (MIRMI), Munich, Germany. 
    E-mails:
    {\tt \{firstname.lastname\}@robotics.utias.utoronto.ca}}%
}
\begin{document}
\maketitle
\thispagestyle{empty}
\pagestyle{empty}

\begin{abstract}
Ultra-wideband (UWB) time difference of arrival (TDOA)-based localization has emerged as a low-cost and scalable indoor positioning solution. However, in cluttered environments, the performance of UWB TDOA-based localization  deteriorates due to the biased and non-Gaussian noise distributions induced by obstacles. In this work, we present a bi-level optimization-based joint localization and noise model learning algorithm to address this problem. In particular, we use a Gaussian mixture model (GMM) to approximate the measurement noise distribution. We explicitly incorporate the estimated state's uncertainty into the GMM noise model learning, referred to as uncertainty-aware GMM, to improve both noise modeling and localization performance. We first evaluate the GMM noise model learning and localization performance in numerous simulation scenarios. We then demonstrate the effectiveness of our algorithm in extensive real-world experiments using two different cluttered environments. We show that our algorithm provides accurate position estimates with low-cost UWB sensors, no prior knowledge about the obstacles in the space, and a significant amount of UWB radios occluded.

\end{abstract}

\input{sections/1-introduction}
\input{sections/2-related_work}

\input{sections/3-problem}

\input{sections/4-method}
\input{sections/5-sim_exp_res}
\input{sections/6-conclusion}

\bibliographystyle{./IEEEtranBST/IEEEtran}
\bibliography{./IEEEtranBST/IEEEabrv,./biblio}

\end{document}

%% file: sections/1-introduction.tex
\section{Introduction}
\label{sec:intro}
Over the last decade, ultra-wideband (UWB) radio technology has been shown to provide high-accuracy time of arrival (TOA) measurements, making it a promising indoor positioning solution. UWB chips have been integrated in the latest generations of consumer electronics including smartphones and smartwatches to support spatially-aware interactions~\cite{uwbNearbyInteration, AndroidUWB}. During the FIFA World Cup 2022, UWB localization technology was used, for the first time, in an official football tournament to enhance the Video Assistant Referee (VAR) system by providing reliable, low-latency, and decimeter-level accurate ball tracking information~\cite{ballSensor, ht-ball}. 

\begin{figure}[t]
    \centering
    \begin{tikzpicture}
    \node[inner sep=0pt] (anchor) at (0.02,2.35)
    {\includegraphics[width=.45\textwidth]{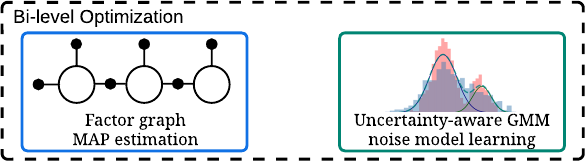}};
    \node[inner sep=0pt, opacity=0.9] (anchor) at (0.2,-1.2)
    {\includegraphics[width=.45\textwidth]{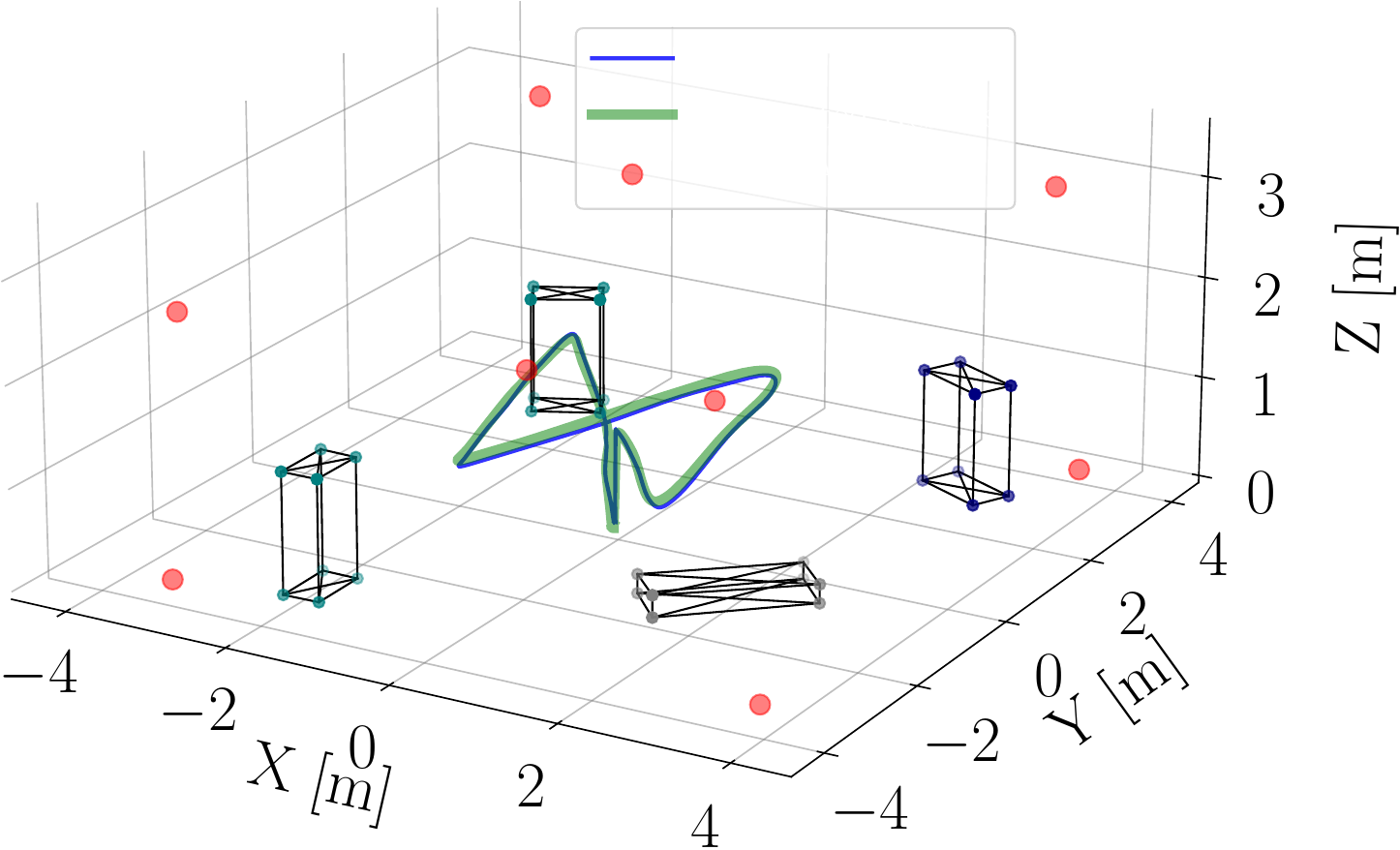}};
    \definecolor{amaranth}{rgb}{0.9, 0.17, 0.31}
    \definecolor{Blue}{rgb}{0.01, 0.28, 1.0}
    \node[text width=1cm, text=black] at (-2.56,2.32) {\small$\mathbf{x}_0$};
    \node[text width=1cm, text=black] at (-1.65,2.32) {\small$\mathbf{x}_1$};
    \node[text width=1cm, text=black] at (-0.72,2.32) {\small$\mathbf{x}_2$};
    \draw [-stealth, line width=1.0pt](-0.5,2.5) -- (0.5,2.5);
    \draw [stealth-, line width=1.0pt](-0.5,2.0) -- (0.5,2.0);
    \node[text width=3cm, text=black] at (1.0, 2.8) {\small$\{\mathcal{X},\bm{\Sigma}\}$};
    \node[text width=3cm, text=black] at (1.45, 1.8) {\small$\bm{\theta}$};
    \node[text width=3cm, text=black] at (-1.4, -1.1) {\scriptsize Wooden Cabinet};
    \node[text width=3cm, text=black] at (0.1, -0.2) {\scriptsize Wooden Cabinet};
    \node[text width=2cm, text=black] at (2.1, -0.65) {\scriptsize Metal Cabinet};
    \node[text width=2cm, text=black] at (0.8, -2.5) {\scriptsize Desk \& Chairs};
    \node[text width=2cm, text=black] at (1.15, 0.9) {\scriptsize Ground Truth};
    \node[text width=2cm, text=black] at (1.15, 0.55) {\scriptsize Proposed Method};
    \node[text width=2cm, text=black] at (1.15, 0.25) {\scriptsize UWB Anchor};
    \draw[Blue, dash pattern=on 1.5pt off 1.5pt] (2.35,2.35) -- (2.8,2.35);
    \node[text width=1cm, text=Blue] at (3.38, 2.55) {\scriptsize noisy};
    \node[text width=1cm, text=Blue] at (3.3, 2.35) {\scriptsize residual};
    \draw[amaranth, dash pattern=on 1.5pt off 1.5pt] (1.65,2.7) -- (2.0, 2.6);
    \node[text width=3cm, text=amaranth] at (2.22, 2.83) {\scriptsize noise-free};
    \node[text width=3cm, text=amaranth] at (2.35, 2.6) {\scriptsize residual};
    \end{tikzpicture}
    \caption{Diagram of the proposed joint localization and noise model learning framework (top) and our localization system setup with one of our experimental results in the first cluttered environment, Env. \#1 (bottom). Our proposed algorithm estimates the robot trajectory through a bi-level optimization framework with the maximum a posteriori (MAP) estimation and uncertainty-aware Gaussian mixture model learning.}
    \label{fig:system-diagram}
\end{figure}
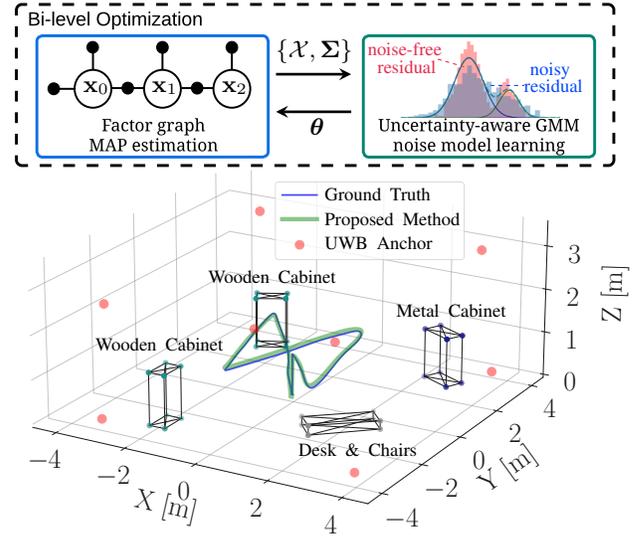

Similar to the Global Positioning System (GPS)~\cite{enge1994global}, an UWB-based positioning system requires UWB radios (also called anchors, see Figure~\ref{fig:system-diagram}) to be pre-installed in the environment as a constellation with known positions, which in turn serve as landmarks for positioning. In  robotics~\cite{pfeiffer2021computationally}, the two main ranging schemes used for UWB localization are \textit{(i)} two-way ranging (TWR) and \textit{(ii)} time difference of arrival (TDOA). In TWR, the UWB module mounted on the robot (also called tag) communicates with an anchor and acquires range measurements through two-way communication. In TDOA, UWB tags compute the difference between the arrival times of the radio packets from two anchors as TDOA measurements. Compared with TWR, TDOA does not require active two-way communication between an anchor and a tag, thus enabling localization of a large number of devices~\cite{hamer2018self}. 

Nonetheless, UWB TDOA-based localization systems still encounter difficulties in cluttered environments. As TDOA localization involves three UWB radios instead of two, the UWB TDOA measurements are easily corrupted by obstacle-induced non-line-of-sight (NLOS) and multi-path propagation in complex environments~\cite{prorok2012online}, leading to degraded positioning accuracy. Recent developments in Gaussian mixture model (GMM)-based residual representation~\cite{pfeifer2019expectation,pfeifer2019incrementally,pfeifer2021advancing} have shown to achieve robust and improved localization performance. However, in these works, the GMM noise model learning process does not account for the noise associated with the measurement residuals induced by the state uncertainty, which can result in inaccurate location estimates when the state is uncertain. 
In this work, we evaluate the uncertainties of measurement residuals through the covariance of the estimated state and leverage this information to improve the GMM noise model learning. Then, we formulate a bi-level optimization to simultaneously perform UWB TDOA localization and uncertainty-aware GMM (U-GMM) noise modeling. We first evaluate the proposed noise model learning and localization performance in numerous simulated problems. Then, we demonstrate the effectiveness of our algorithm with extensive real-world experiments using low-cost UWB sensors in two different cluttered environments. We show that our algorithm, compared to conventional methods, provides improved position estimates without prior knowledge about the obstacles in the space and a significant amount of UWB radios occluded. 
%
%
Our main contributions can be summarized as follows:
\begin{enumerate}
  \item We explicitly leverage the uncertainty of the estimated state to improve the GMM noise model learning performance. 
  
  \item We present a bi-level optimization framework for joint localization and uncertainty-aware noise model learning to improve UWB TDOA localization performance.
   
  \item We evaluate the proposed noise model learning and localization performance in numerous simulated problems. We further demonstrate the effectiveness of our proposed method in extensive real-world experiments using two different cluttered environments.
  
\end{enumerate}

%% file: sections/2-related_work.tex
\section{Related Work}
\label{sec:related_work}



Multiple approaches have been proposed to improve UWB localization performance with biased and non-Gaussian measurement noises. Gaussian processes~\cite{ledergerber2017ultra} and neural networks~\cite{zhao2021learning} trained with the ground truth measurements from a motion capture system have been used to model the UWB measurement biases. The proposed method in~\cite{ledergerber2018calibrating} learns a Gaussian process noise model without ground truth information, but still requires training data to learn the model beforehand. The iterative measurement bias learning approach proposed in~\cite{van2020iterative} does not have an offline training step, but requires repeating the same trajectory multiple times to learn the bias model.
%
More generic mechanisms to reduce the influence of the biases and measurement outliers are M-estimators~\cite{mactavish2015all}, which apply robust cost functions to downweight large measurement residuals. For biased and asymmetric non-Gaussian noise distributions, the performance of M-estimators could deteriorate due to the symmetric nature of the robust cost functions\cite{pfeifer2021advancing}. 

More recently, authors in~\cite{pfeifer2019expectation, pfeifer2019incrementally} leverage Gaussian mixture models (GMM) to represent the distribution of measurement residuals and improve the localization performance. GMMs are flexible enough to represent asymmetric, multimodal, and skewed measurement noise distributions, which also can be used for efficient nonlinear least squares optimization~\cite{rosen2013robust,pfeifer2021advancing}. 
%
In the context of state estimation, the measurement residuals, which are computed based on the estimated state and used for GMM noise modeling, are often associated with uncertainties. However, the standard GMM noise model learning approaches in~\cite{pfeifer2019expectation, pfeifer2019incrementally} disregard the uncertainties associated with the residuals, which can result in degraded GMM noise modeling performance. Consequently, it is necessary to leverage the uncertainty of the estimated state to jointly improve the GMM noise modeling and localization performance.

In this work, we present a bi-level optimization-based joint localization and uncertainty-aware GMM noise model learning algorithm, which explicitly leverages the uncertainty of the estimated state to improve the noise modeling and localization performance. We demonstrate the effectiveness of the proposed method in numerous simulated problems and real-world experiments. To the best of our knowledge, this is the first work to incorporate the uncertainty of the estimated state into GMM noise model learning to improve the accuracy and robustness of UWB TDOA localization.

%% file: sections/3-problem.tex
\section{Problem Formulation}
\label{sec:problem}
We consider an UWB TDOA localization system in a cluttered indoor environment. The set of $m_a$ UWB anchors are divided into TDOA anchor pairs $\Gamma = \{(1,2), \cdots, (m_a-1, m_a)\}$ and are assumed to be fixed in the space $\mathcal{P} \in \mathbb{R}^{n_d}$ with $n_d=\{1,2,3\}$ indicating the dimension. To facilitate our analysis, we define a vector $\bm{a}=\left[\bm{a}_1^T, \bm{a}_2^T, \cdots, \bm{a}_{m_a}^T\right]^T \in \mathbb{R}^{n_d\cdot m_a}$ that contains all anchor positions. The general state to be estimated is represented as $\mathcal{X}$, which contains the robot's poses along the trajectory. We refer to the absolute coordinate frame created by the UWB anchors as the inertial frame~$\mathcal{F}_{\mathcal{I}}$ and denote the robot body frame as~$\mathcal{F}_{\mathcal{B}}$.  

Our problem is described as follows. The robot is assumed to be equipped with an UWB tag together with a source of odometry providing incremental motion information. We indicate the odometry measurements as $\mathcal{U}$ and the noisy UWB TDOA measurements as $\mathcal{D}$. In cluttered environments, UWB measurements are often affected by obstacles, leading to biased and non-Gaussian measurement errors. We use Gaussian mixture models (GMMs), parameterized by $\bm{\theta}$, to approximate the UWB error distributions. Our goal is to estimate the state $\mathcal{X}$ and the noise parameters $\bm{\theta}$ through maximizing the joint likelihood: 
\begin{equation}
    \mathcal{X}^{\star}, \bm{\theta}^{\star} = \argmax_{\mathcal{X}, \bm{\theta}}~p\left(\mathcal{X}, \mathcal{U}, \mathcal{D} |\bm{\theta}\right).
\end{equation}

%% file: sections/4-method.tex
\section{Methodology}
%

Gaussian mixture models (GMMs) are well-suited to representing the biased and non-Gaussian UWB TDOA measurement noises in cluttered environments, which enable accurate and robust localization performance. However, the conventional GMM noise model learning often ignores the uncertainties of the measurement residuals, which deteriorates the modeling performance. In this section, we incorporate the uncertainty of the estimated state into measurement noise model learning to jointly improve the noise modeling and localization performance.

\subsection{UWB TDOA Localization via Maximum a Posteriori}
\label{subsec:uwb_local}
Considering a general UWB TDOA-based localization system, the noisy UWB TDOA measurement between the robot pose $\mathbf{x}_{t_n} \in \mathcal{X}$ at discrete time $t_n$  and anchor pair $\{\bm{a}_i, \bm{a}_j\}$ is modeled as
\begin{equation}
\small
    \begin{split}
    d_{ij,t_n} &= \bar{d}_j(\mathbf{x}_{t_n}) - \bar{d}_i(\mathbf{x}_{t_n}) + \eta_{ij,t_n}   \\
    & = \bar{d}_{ij}(\mathbf{x}_{t_n}) + \eta_{ij,t_n},
    \end{split}
\end{equation}
where $\bar{d}_i(\mathbf{x}_{t_n})$ and $\bar{d}_j(\mathbf{x}_{t_n})$ are the error-free range measurements between the robot pose $\mathbf{x}_{t_n}$ and anchor positions $\{\bm{a}_i, \bm{a}_j\}, (i,j)\in \Gamma$ and $\eta_{ij,t_n}$ is the measurement error. 

We denote the set of odometry measurements as $\mathcal{U}=\{\mathbf{u}_t\}$ at discrete times $t=1, \cdots, T$ and the set of TDOA measurements from anchor pair $\{\bm{a}_i, \bm{a}_j\}$ as $\mathcal{D}_{ij}=\{d_{ij,t_1}, \cdots, d_{ij,t_N}\}$. We summarize all the observed UWB measurements as $\mathcal{D} = \{\mathcal{D}_{ij}\}, (i,j)\in \Gamma$. Assuming odometry and UWB TDOA measurements are independent, the joint likelihood of the robot poses and the observations is
\begin{equation}
\small
    p(\mathcal{X}, \mathcal{U}, \mathcal{D}) = p(\mathbf{x}_0)\prod_{t=1}^T p(\mathbf{x}_t|\mathbf{x}_{t-1},\mathbf{u}_{t})\prod_{(i,j)\in \Gamma} \prod_{n=1}^N p(d_{ij,t_n}|\mathbf{x}_{t_n}),
\end{equation}
where $p(\mathbf{x}_0)$ is a prior on the initial state, $p(\mathbf{x}_t|\mathbf{x}_{t-1},\mathbf{u}_{t})$ is the \textit{motion model}, and $p(d_{ij,t_n}|\mathbf{x}_{t_n})$ is the UWB TDOA \textit{measurement model}. The UWB TDOA-based localization problem corresponds to maximizing the joint likelihood
\begin{equation}
\label{eq:map}
    \hat{\mathcal{X}} = \argmax_{\mathcal{X}}~p(\mathcal{X},\mathcal{U},\mathcal{D}),
\end{equation}
leading to a \textit{Maximum a Posteriori (MAP)} estimation result.

\subsection{Gaussian Mixture Model for Nonlinear Least Squares}
\label{subsec:gmm}
\begin{figure}[t]
    \center
    \begin{tikzpicture}[
    node distance={9mm},
    vertex/.style={circle, draw=black!100, thick, minimum width=0.85cm},
    empty_factor/.style={minimum size=1mm},
    ]
    \node[circle,fill=black,inner sep=0pt,minimum size=5pt,label=below:{$\phi_{0}$}] (prior) at (0,0) {};
    \node[vertex] (x0)  [right of=prior] {$\mathbf{x}_{0}$};  
    \node[circle, draw, right of=x0, fill=black, inner sep=0pt,minimum size=5pt, label=below:{$\phi_{\mathbf{u}_1}$}] (o0){};
    \node[vertex] (x1)  [right of=o0] {$\mathbf{x}_{1}$};
    \node[empty_factor] (o1) [right of=x1] {$\hdots$};
    \node[vertex] (xM) [right of=o1] {$\mathbf{x}_{t_n}$};
    \node[circle, draw, above of=x0, fill=black, inner sep=0pt,minimum size=5pt, label=above:{$\phi_{d_{12,0}|\bm{\theta}_{12}}$}] (y0){};
    \node[circle, draw, above of=x1, fill=black, inner sep=0pt,minimum size=5pt, label=above:{$\phi_{d_{23,1}|\bm{\theta}_{23}}$}] (y1){};
    \node[circle, draw, above of=xM, fill=black, inner sep=0pt,minimum size=5pt, label=above:{$\phi_{d_{ij,t_n}|\bm{\theta}_{ij}}$}] (yM){};
    %
    
    \draw (prior) -- (x0);
    \draw (x0) -- (o0);
    \draw (o0) -- (x1);
    \draw (x1) -- (o1);
    \draw (o1) -- (xM);
    \draw (x0) -- (y0);
    \draw (x1) -- (y1);
    \draw (xM) -- (yM);
    \end{tikzpicture}
        \caption{The factor graph for the GMM-based UWB localization. The prior estimate of the initial state is added as an unary factor $\phi_{0}$ to the graph. Odometry binary factors are indicated as $\phi_{\mathbf{u}_t}$ and the UWB TDOA unary factors $\phi_{d_{ij,t_n}|\bm{\theta}_{ij}}$ are parameterized by the GMM hyperparameter $\bm{\theta}_{ij}$.}
    \label{fig:factor_graph}
\end{figure}
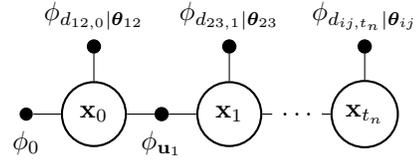

In cluttered environments, the UWB measurement errors $\{\eta_{ij,t_n}\}$ often show biased and non-Gaussian distributions due to degraded radio signals caused by NLOS and multi-path radio propagation~\cite{van2020iterative,prorok2012online}. We use Gaussian mixture models (GMMs) to model those distributions due to their flexibility. At each time step $t_n$, we compute the TDOA measurement residual based on the estimated state $\hat{\mathbf{x}}_{t_n}$ and the observed TDOA values $d_{ij,t_n}$ as
\begin{equation}
\label{eq:residual_model}
    r_{ij,t_n}(d_{ij,t_n},\hat{\mathbf{x}}_{t_n}) = d_{ij,t_n} -\bar{d}_{ij}(\hat{\mathbf{x}}_{t_n}).   
\end{equation}
For the TDOA measurements $\mathcal{D}_{ij}$, we use a GMM with $K$ Gaussian components parameterized by the hyperparameter $\bm{\theta}_{ij}=[\pi_{ij}^{1}, \cdots,\pi_{ij}^{K},\mu_{ij}^{1},\cdots,\mu_{ij}^{K},\sigma_{ij}^1,\cdots,\sigma_{ij}^K]$ to model the distribution of the measurement noise $\eta_{ij,t_n},(i,j)\in \Gamma$. The parameter $\{\pi_{ij}^{k},\mu_{ij}^{k},\sigma_{ij}^k\}$ indicates the weight, mean, and standard deviation of the $k$-th Gaussian component, respectively. The set of hyperparameters is denoted as $\bm{\theta}=\{\bm{\theta}_{ij}\}$. The factor graph for the GMM-based UWB localization is shown in Figure~\ref{fig:factor_graph}. We indicate the odometry binary factor as $\phi_{\mathbf{u}_t}$ and the UWB TDOA unary factor parameterized by the hyperparameter $\bm{\theta}_{ij}$ as $\phi_{d_{ij,t_n}|\bm{\theta}_{ij}}$.

We apply the Max-Sum-Mixture approach proposed in~\cite{pfeifer2021advancing} to convert a GMM-based maximum likelihood estimation into a nonlinear least squares optimization. With the GMM hyperparameter $\bm{\theta}_{ij}$, the likelihood of $d_{ij,t_n}$ has the following relationship
\begin{equation}
    p(d_{ij,t_n}|\hat{\mathbf{x}}_{t_n}, \bm{\theta}_{ij}) \propto \sum_{k=1}^{K} s_k\exp{(e_k(r_{ij,t_n}))},
\end{equation}
where $s_k=\frac{\pi_{ij}^k}{\sigma_{ij}^k}$, {\footnotesize$e_k(r_{ij,t_n}) = -\frac{1}{2}\left(\frac{r_{ij,t_n} - \mu_{ij}^k}{\sigma_{ij}^k}\right)^2$}, and we drop the functional dependencies in Equation~\eqref{eq:residual_model} and indicate the residual as $r_{ij,t_n}$ for short. Following the derivation in~\cite{pfeifer2021advancing}, we define $\tilde{k} = \argmax_k s_k\exp{(e_k(r_{ij,t_n}))}$, indicating the index of the dominant Gaussian mode, and compute the square root of the cost function $\bm{\rho}(r_{ij,t_n})$ as:
\begin{equation}
\footnotesize
    \bm{\rho}(r_{ij,t_n}) = \begin{bmatrix} \displaystyle \frac{r_{ij,t_n}-\mu_{ij}^{\tilde{k}}}{\sigma_{ij}^{\tilde{k}}} \\  
    \addlinespace
\sqrt{-2\ln{\left(\displaystyle\frac{1}{\zeta}\sum\limits_{k=1}^K s_k\exp{(e_k(r_{ij,t_n})-e_{\tilde{k}}(r_{ij,t_n}))}\right )}}
\end{bmatrix}, 
\end{equation}
where $\zeta = K \cdot \max_k(s_k) + c$ is the normalization constant that guarantees the expression inside the square root to be positive and $c$ is set to be $10$ following the suggestion in~\cite{pfeifer2021advancing}.

Now, we formulate the nonlinear least squares problem. We denote a general motion model as 
\begin{equation}
    \mathbf{x}_t = f(\mathbf{x}_{t-1}, \mathbf{u}_{t}) + \mathbf{w}_{t},
\end{equation}
where $\mathbf{w}_t \sim \mathcal{N}(\mathbf{0}, \bm{\Sigma}_{\mathbf{u}})$ is the additive white Gaussian noise (AWGN). Denoting the prior estimate for the initial state as $\mathbf{x}_0 \sim \mathcal{N}(\check{\mathbf{x}}_0, \check{\bm{\Sigma}}_{0})$, we convert the UWB localization problem with GMM measurement noises into a nonlinear least square problem as follows:
\begin{equation}
\small
\begin{split}
\hat{\mathcal{X}} &= \argmax_{\mathcal{X}}~p(\mathcal{X},\mathcal{U},\mathcal{D}|\bm{\theta}) \\
&= \argmin_{\mathcal{X}}\left\{\|\check{\mathbf{x}}_0-\mathbf{x}_0\|^2_{\check{\bm{\Sigma}}_0} + \sum_{t=1}^{T}\|f(\mathbf{x}_{t-1}, \mathbf{u}_{t})-\mathbf{x}_t\|^2_{\bm{\Sigma}_\mathbf{u}} \right. \\
& ~~~~~~~~~~~~ + \left.\sum_{(i,j)\in\Gamma}\sum_{n=1}^N \|\bm{\rho}(r_{ij,t_n})\|^2 \right\},
\end{split}
\label{eq:map_nls}
\end{equation}
where $\|\cdot\|^2_{\bm{\Sigma}_n}$ is the squared Mahalanobis distance given a covariance matrix $\bm{\Sigma}_n$ and $\|\bm{\rho}(r_{ij,t_n})\|^2=\bm{\rho}(r_{ij,t_n})^T\bm{\rho}(r_{ij,t_n})$.

\subsection{Uncertainty-aware GMM Noise Model Learning using Variational Inference}
\label{sec:vi}
%
Conventional GMM noise model learning methods~\cite{pfeifer2019expectation,pfeifer2019incrementally} often disregard the uncertainties of the measurement residuals, resulting in degraded modeling performance. In the context of state estimation, we are able to quantify the uncertainties of the measurement residuals using the covariance of the estimated state. To improve the measurement noise model learning performance, we explicitly incorporate the estimated state's uncertainty into the GMM noise model learning using a variational inference framework. In the proposed uncertainty-aware GMM noise model learning method, we wish to estimate the hyperparameter $\bm{\theta}_{ij}$ of the GMM noise model $\eta_{ij,t_n}\sim \textrm{GMM}(\bm{\theta}_{ij})$ through a set of noisy measurement residuals. For ease of notation, we consider a general GMM noise model learning and drop the anchor pair subscript $ij$. Moreover, we simplify the time step subscript from $t_n$ to $n$ for brevity. We indicate the set of $N$ observed measurement residuals as $\mathcal{R}=\{r_{1}, \cdots, r_{N}\}$. The noise-free measurement residuals $\mathcal{E}=\{\eta_{1}, \cdots, \eta_{N}\}$ are the corresponding latent variables. We indicate the estimated state at time $t_{n}$ as $\mathbf{x}_{n} \sim \mathcal{N}(\hat{\mathbf{x}}_{n},\hat{\bm{\Sigma}}_{n})$ where $\hat{\bm{\Sigma}}_{n}$ is the covariance matrix of the state. According to Equation \eqref{eq:residual_model}, the measurement residual $r_{n}$ is computed by the estimated state $\hat{\mathbf{x}}_{n}$ and the observed TDOA measurement $d_{n}$. We propagate the covariance matrices of the estimated state $ \hat{\bm{\Sigma}} = \{\hat{\bm{\Sigma}}_{1},\cdots,\hat{\bm{\Sigma}}_{N}\}$ through the nonlinear residual model to approximate the residuals' uncertainties $\Phi = \{\varphi_{1}, \cdots,\varphi_{N}\}$ using the sigma-point transformation~\cite{julier1996general}. Hence, each observed residual can be modeled as a noisy sample drawn from a Gaussian distribution $\mathcal{N}(r_{n}|\eta_{n}, \varphi_{n})$, where the mean is the error-free residual value $\eta_{n}$ and the variance $\varphi_{n}$ represents the sample uncertainty. 


We use a Gaussian mixture model with $K$ Gaussian components to represent the latent noise-free measurement residuals $\mathcal{E}=\{\eta_{1}, \cdots, \eta_{N}\}$. Under a latent variable model, we associate each residual $r_{n}$ with a binary latent variable $\mathbf{z}_n =[z_{n1}, \cdots, z_{nK}]^T$, in which only one element is set to one to indicate that the true residual $\eta_{n}$ was generated from that Gaussian component. We indicate the latent variables as $\mathcal{Z}=\{\mathbf{z}_1, \cdots,\mathbf{z}_N\}$ and the sets of the GMM parameters as $\Pi = \{\pi_1,\cdots, \pi_K\}$, $\mathcal{M}=\{\mu_1, \cdots, \mu_K\}$, and $\Lambda = \{\lambda_1,\cdots, \lambda_K\}$, where $\lambda_k=\frac{1}{\sigma_k^2}$ is the precision of the $k-$th Gaussian component.Therefore, we follow a similar approach in~\cite{hou2008robust} and have the following probabilistic models
\begin{equation}
\begin{split}
    & p(\mathcal{R}|\mathcal{E}, \Phi) = \prod_{n=1}^N \mathcal{N}(r_{n}|\eta_{n},\varphi_n), ~~ p(\mathcal{Z}|\Pi) = \prod_{n=1}^N \prod_{k=1}^K \pi_{k}^{z_{nk}}, \\
    &p(\mathcal{E}|\mathcal{Z},\mathcal{M},\Lambda) = \prod_{n=1}^N \prod_{k=1}^K \mathcal{N}(\eta_{n}|\mu_k, \lambda_k^{-1})^{z_{nk}}. 
\end{split}
\end{equation}

We follow~\cite{bishop2006pattern} and place a Dirichlet distribution and a Gaussian-Wishart distribution as the prior distributions for $p(\Pi)$ and $p(\mathcal{M}, \Lambda)$
\begin{equation}
\begin{split}
    p(\Pi) &= \mathrm{Dir}(\Pi|\bm{\alpha}_0) = C(\bm{\alpha}_0)\prod_{k=1}^K\pi_k^{\alpha_0-1} \\
    p(\mathcal{M}, \Lambda) &= \prod_{k=1}^K \mathcal{N}(\mu_k|m_0, (\beta_0\lambda_k)^{-1})\mathcal{W}(\lambda_k| w_0, \nu_0),
\end{split}
\end{equation}
where $C(\bm{\alpha}_0)$ is the normalization constant for the Dirichlet distribution with parameter $\bm{\alpha}_0$, $\mathcal{N}(\mu_k|m_0, (\beta_0\lambda_k)^{-1})$ is the Gaussian distribution with mean $m_0$ and precision $\beta_0\lambda_k$, and $\mathcal{W}(\lambda_k| w_0, \nu_0)$ is the Wishart distribution with scale variable $w_0$ and degrees of freedom $\nu_0$. We assume the set of residuals' uncertainties $\Phi$ is observed through uncertainty transformation and the joint probability distribution can be factorized as 
\begin{equation}
\small
\begin{split}
    p(\mathcal{R}, \mathcal{E}, \mathcal{Z}, \mathcal{M}, \Lambda, \Pi | \Phi) = &p(\mathcal{R} | \mathcal{E}, \Phi) p(\mathcal{E}|\mathcal{Z}, \mathcal{M}, \Lambda)p(\mathcal{Z}|\Pi) \\
    &p(\Pi)p(\mathcal{M}|\Lambda)p(\Lambda).
\end{split}
\end{equation}
We provide a detailed derivation of the uncertainty-aware GMM parameter learning in the supplementary document\footnote{\hypertarget{fn:supplementary}{\url{http://tiny.cc/iros23_supplementary_doc}}}. Readers may also refer to \cite{hou2008robust, bishop2006pattern} for further information.

In variational inference, we approximate the posterior distribution of the latent variable with a variational distribution, which is commonly chosen through minimizing the Kullback-Leibler (KL) divergence. In our probabilistic model, we denote the latent variables as $\mathcal{H} = \{\mathcal{E},\mathcal{Z}, \Pi, \mathcal{M},\Lambda\}$ and the KL divergence is computed as 
\begin{equation}
    \textrm{KL}(q||p) = -\int q(\mathcal{H})\ln{\left(\frac{p(\mathcal{H}|\mathcal{R},\Phi)}{q(\mathcal{H})}\right)} d\mathcal{H}.
\end{equation}
Based on the mean-field-approximation, we assume the variational distribution can be factorized as 
\begin{equation}
    q(\mathcal{H})=q(\mathcal{E}, \mathcal{Z}, \Pi, \mathcal{M}, \Lambda) = q(\mathcal{E}| \mathcal{Z}) q(\mathcal{Z}) q(\Pi, \mathcal{M}, \Lambda),
\end{equation}
which can be further factorized as
\begin{equation}
    \begin{split}
        q(\mathcal{E}|\mathcal{Z}) &= \prod_{n=1}^N \prod_{k=1}^K q(\eta_n|z_{nk}=1)^{z_{nk}}  \\
        q(\Pi,\mathcal{M},\Lambda) &= q(\Pi)\prod_{k=1}^K q(\mu_k|\lambda_k)q(\lambda_k).
    \end{split}
\end{equation}
We apply the general expression in variational inference~\cite{bishop2006pattern} to compute the optimal solution of $q(\mathcal{E}| \mathcal{Z})$, $q(\mathcal{Z})$, and $q(\Pi, \mathcal{M}, \Lambda)$. Based on the general expression, we have
\begin{equation}
\small
\begin{split}
    \ln{q^{\star}(\mathcal{Z},\mathcal{E})} &= \underset{\Pi, \mathcal{M}, \Lambda}{\mathbb{E}}\left[\ln{p(\mathcal{R}, \mathcal{E}, \mathcal{Z}, \mathcal{M}, \Lambda, \Pi | \Phi)}\right] + \textrm{const}   \\
    \ln{q^{\star}(\mathcal{Z})} &= \underset{\mathcal{E}}{\mathbb{E}}\left[\ln{q^{\star}(\mathcal{Z},\mathcal{E})}\right] - \underset{\mathcal{E}}{ \mathbb{E}}\left[\ln{q^{\star}(\mathcal{E} | \mathcal{Z})}\right].
\end{split}
\end{equation}
The constant term on the right-hand side indicates terms independent of the latent variable $\mathcal{Z}$ and $\mathcal{E}$. 
Since we select the conjugate prior distribution for $q(\mu_k,\lambda_k)$, the optimal variational distribution has the same Gaussian-Wishart form, indicated as $q(\mu_k,\lambda_k) = \mathcal{N}(\mu_k|m_k, (\beta_k\lambda_k)^{-1})\mathcal{W}(\lambda_k| w_k, \nu_k)$. Therefore, we have the following expression
\begin{equation}
\footnotesize
\label{eq:ln_q_ez}
\begin{split}
    &\ln{q^{\star}(\mathcal{E} | \mathcal{Z})} = \ln{q^{\star}(\mathcal{Z},\mathcal{E})} - \ln{q^{\star}(\mathcal{Z})} \\
    =& \sum_{n=1}^{N}\sum_{k=1}^{K}z_{nk}\left(\ln{\mathcal{N}(r_n|\eta_n,\varphi_n)}+\ln{\mathcal{N}(\eta_n|m_k, (\nu_k w_k)^{-1})}\right) + \textrm{const}.
\end{split}
\end{equation}
and the optimal solution $q^{\star}(\eta_n|z_{nk}=1)$ is a Gaussian distribution~\cite{hou2008robust} 
\begin{equation}
    q^{\star}(\eta_n|z_{nk}=1) = \mathcal{N}(\eta_{n|k}|\tau_{n|k}, \psi_{n|k}),
\end{equation}
where the mean and the variance are computed as:
\begin{equation}
\tau_{n|k} = \psi_{n|k}\left(\frac{r_n}{\varphi_n} + \nu_k w_k m_k\right), 
\psi_{n|k} = \frac{\varphi_n}{1 + \varphi_n \nu_k w_k}.
\end{equation}
We make use of the expression of Equation~\eqref{eq:ln_q_ez} to compute the optimal solution for $q^{\star}(\mathcal{Z})$
\begin{equation}
\small
    q^{\star}(\mathcal{Z}) = \prod_{n=1}^N \prod_{k=1}^K \gamma_{nk}^{z_{nk}}, ~~
    \gamma_{nk} = \frac{\varrho_{nk}}{\sum_{l=1}^K \varrho_{nl}} = \mathbb{E}[z_{nk}],
\end{equation}
where $\gamma_{nk}$ is the responsibility of the $k$-th component for the $n$-th residual sample, and $\varrho_{nk}$ can be computed as 
\begin{equation}
\small
\begin{split}
    \ln{\varrho_{nk}} = & \underset{\mathcal{E}}{\mathbb{E}}\left[\ln{\mathcal{N}(r_n|\eta_n, \varphi_n)}\right] + \underset{\mathcal{E},\mathcal{M},\Lambda}{\mathbb{E}}\left[\ln{\mathcal{N}(\eta_n|\mu_k, \lambda_k^{-1})}\right]   \\ 
    & + \underset{\Pi}{\mathbb{E}}\left[\ln{\pi_k} \right] - \underset{\mathcal{E}}{\mathbb{E}}\left[\ln{\mathcal{N}(\eta_n| \tau_{n|k}, \psi_{n|k})}\right].    
\end{split}
\end{equation}
The optimal solutions for $q^{\star}(\Pi)$ and $q^{\star}(\mathcal{M},\lambda)$ are similar to the derivation in~\cite{hou2008robust} and \cite{bishop2006pattern} and we only provide the final expressions in here:
\begin{equation}
\small
\begin{split}
    q^{\star}(\Pi) &= \textrm{Dir}(\Pi|\bm{\alpha})  \\
    q^{\star}(\mathcal{M},\Lambda) &= \prod_{k=1}^K \mathcal{N}(\mu_k|m_k, (\beta_k\lambda_k)^{-1})\mathcal{W}(\lambda_k| w_k, \nu_k),
\end{split}
\end{equation}
where $\bm{\alpha}= [\alpha_1, \cdots, \alpha_K]$ with $\alpha_k = \alpha_0 + N_k$, $N_k=\sum_{n=1}^{N}\gamma_{nk}$ and the Gaussian-Wishart parameters are computed as follows:
\begin{equation}
\small
\begin{split}
    \beta_k &= \beta_0 + N_k, ~~~~~~~~~~~~~~~~~    m_k =\frac{1}{\beta_k}(\beta_0 m_0+N_k\bar{\tau}_k),  \\
    \bar{\tau}_k &= \frac{1}{N_k}\sum_{n=1}^N\underset{\mathbf{z}_n}{\mathbb{E}}[z_{nk}]\tau_{n|k},  ~~~~~    \nu_k = \nu_0 + N_k, \\
    w_k^{-1} &= w_0^{-1} + N_k S_k + \frac{\beta_0 N_k}{\beta_0 + N_k}(\bar{\tau}_k - m_0)^2, \\
    S_k &= \frac{1}{N_k}\sum_{n=1}^N\underset{\mathbf{z}_n}{\mathbb{E}}[z_{nk}](\tau_{n|k}-\bar{\tau}_k)^2 + \frac{1}{N_k}\sum_{n=1}^N\underset{\mathbf{z}_n}{\mathbb{E}}[z_{nk}]\psi_{n|k}.
\end{split}
\end{equation}

The optimization of the variational distributions involves alternating between the variational E and M steps. In the variational E step, we evaluate the responsibilities $\mathbb{E}[z_{nk}]$. In the variational M step, we keep the responsibilities fixed and use them to compute the variational distributions $q^{\star}(\mathcal{E}|\mathcal{Z})$, $q^{\star}(\Pi)$, and $q^{\star}(\mathcal{M},\Lambda)$. The convergence of the iterative process is monitored by computing the variational evidence lower bound (ELBO)
\begin{equation}
\mathcal{L}(q) = \int q^{\star}(\mathcal{H})\ln{\left(\frac{p(\mathcal{H},\mathcal{R}|\Phi)}{q^{\star}(\mathcal{H})}\right)} d\mathcal{H}, 
\label{eq:elbo}
\end{equation}
which should monotonically increase after each iteration.

\subsection{Bi-level Optimization for Joint Localization and Noise Model Learning}
To jointly solve the UWB localization problem and learn the GMM noise parameters, we propose a bi-level optimization algorithm~\cite{pineda2022theseus} shown as follows:  
\begin{equation}
    \begin{split}
       \textrm{Inner loop} &:  ~~~~ \hat{\mathcal{X}} = \argmax_{\mathcal{X}} p(\mathcal{X},\mathcal{U},\mathcal{D}|\bm{\theta}), \\
       \textrm{Outer loop} &:  ~\hat{q}(\bm{\theta}) = \argmax_{q(\bm{\theta})} \mathcal{L}(q(\bm{\theta})|\mathcal{X},\bm{\Sigma}),  \\
       & ~~~~~~~ \hat{\bm{\theta}} = \underset{\bm{\theta}}{\mathbb{E}}[\hat{q}(\bm{\theta})],
    \end{split}
    \label{eq:bi-level}
\end{equation}
where we introduce the functional dependencies of $\bm{\theta}$, $\mathcal{X}$, and $\bm{\Sigma}$ to the outer loop optimization for clarity.

In the inner loop optimization, we solve the factor graph-based least squares optimization in Equation~\eqref{eq:map_nls} and achieve the MAP estimate $\hat{\mathcal{X}}$ based on the current noise parameter $\bm{\theta}$. The covariance $\hat{\bm{\Sigma}}$ is then built from the curvature of the log-posterior at the maximum~\cite{opper2009variational}, which is also known as the Laplace approximation~\cite{bishop2006pattern}. In the outer loop optimization, we optimize the variational distribution $q(\bm{\theta})=q(\mathcal{Z},\mathcal{M},\Lambda)$ by maximizing the ELBO in Equation~\eqref{eq:elbo}. The GMM parameters $\bm{\theta}$ are obtained through the expectation over the estimated variational distribution $\hat{q}(\bm{\theta})$. The algorithm can be initialized by an odometry dead-reckoning, conventional MAP estimations with Gaussian assumptions, or other methods that can provide an initial estimate of $\{\mathcal{X}, \bm{\Sigma}\}$.


We perform the bi-level optimization iteratively and check the convergence based on the loss value of the inner loop cost function in Equation~\eqref{eq:map_nls}. The designed bi-level optimization algorithm aims to find the state and noise parameter that maximize the joint likelihood of the state and observations. The algorithm terminates when the loss value increases, which indicates the joint likelihood decreases, or the maximum number of iterations is reached. 


%% file: sections/5-sim_exp_res.tex
\section{Simulation and Experimental Results}
\label{sec:sim-exp}
In this section, we present simulation and experimental results of the proposed joint localization and uncertainty-aware noise model learning algorithm. In all the simulations and experiments, we use Gaussian mixture models with three Gaussian components ($K=3$), as it is typically sufficient for UWB noise modeling~\cite{prorok2012online,pfeifer2019incrementally}. We first demonstrate the performance of the proposed algorithm on simulated 1-D, 2-D, and 3-D TDOA localization problems. Then, we evaluate the uncertainty-aware GMM (U-GMM) model learning performance under different uncertainty levels with simulated data. Finally, we conducted extensive real-world experiments in two different cluttered indoor environments for 3-D pose estimation. We demonstrate the effectiveness of the proposed method through the improvement of localization performance compared to conventional methods. Readers may refer to the supplementary material\hyperlink{fn:supplementary}{$^1$} for detailed descriptions of the methods to which we compared and the hyperparameters used in all the simulations and experiments.

\subsection{Simulation Results on Localization Problems}
As our proposed method is a general localization framework, we validate the localization accuracy and GMM model learning performance in simple 1-D and 2-D simulated problems as well as a more realistic 3-D pose estimation problem. In the 1-D simulation setup, we consider a TDOA localization problem for a mobile robot moving along a line with one pair of anchors $\{a_1, a_2\}$. We assume the robot state $x_t$ is between the anchors with $x_t \in (a_1, a_2)$ to obtain the following linear motion and measurement models:
\begin{equation}
\small
\begin{split}
    \textrm{motion model :}~~ & x_t = x_{t-1} + u_{t} + w_t \\
    \textrm{meas. model :}~~  & d_{12,t} = (a_1 + a_2) - 2 x_t + \eta_{12,t}, 
\end{split}
\end{equation}
where $w_t \sim \mathcal{N}(0,\sigma_u^2)$ is the noise of the odometry measurement $u_t$ and $\eta_{12,t}$ is the simulated GMM measurement noise.  In the 2-D simulation setup, we consider the robot state as $\mathbf{x}_t = [x_t,y_t]^T$ and two pairs of anchors $\Gamma=\{(1,2),(3,4)\}$ for localization. We keep a linear motion model but the TDOA hyperbolic measurement model is nonlinear:
\begin{equation}
\small
\begin{split}
    \textrm{motion model :}~~ & \mathbf{x}_t = \mathbf{x}_{t-1} + \mathbf{u}_{t} + \mathbf{w}_t \\
    \textrm{meas. model :}~~  & d_{ij,t_n} =  \|\mathbf{x}_{t_n} - \bm{a}_j\| - \|\mathbf{x}_{t_n} - \bm{a}_i\| + \eta_{ij,t_n}, 
\end{split}
\end{equation}
where $\|\cdot\|$ is the $\ell_2$ norm, $\mathbf{w}_{t} \sim \mathcal{N}(\bm{0}, \bm{\Sigma}_{\mathbf{u}})$ is the noise of the odometry measurement $\mathbf{u}_{t} = [\Delta x_{t}, \Delta y_{t}]^{T}$ and $\eta_{ij,t_n}$ is the simulated GMM noise. In both 1-D and 2-D simulations, the GMM noise distributions are simulated as mixture models with two Gaussian components.
%
\begin{figure}[tb]
    \centering
    \begin{tikzpicture}
    \node[inner sep=0pt, opacity=1.0] (anchor) at (0.0,0.0)
    {\includegraphics[width=.49\textwidth]{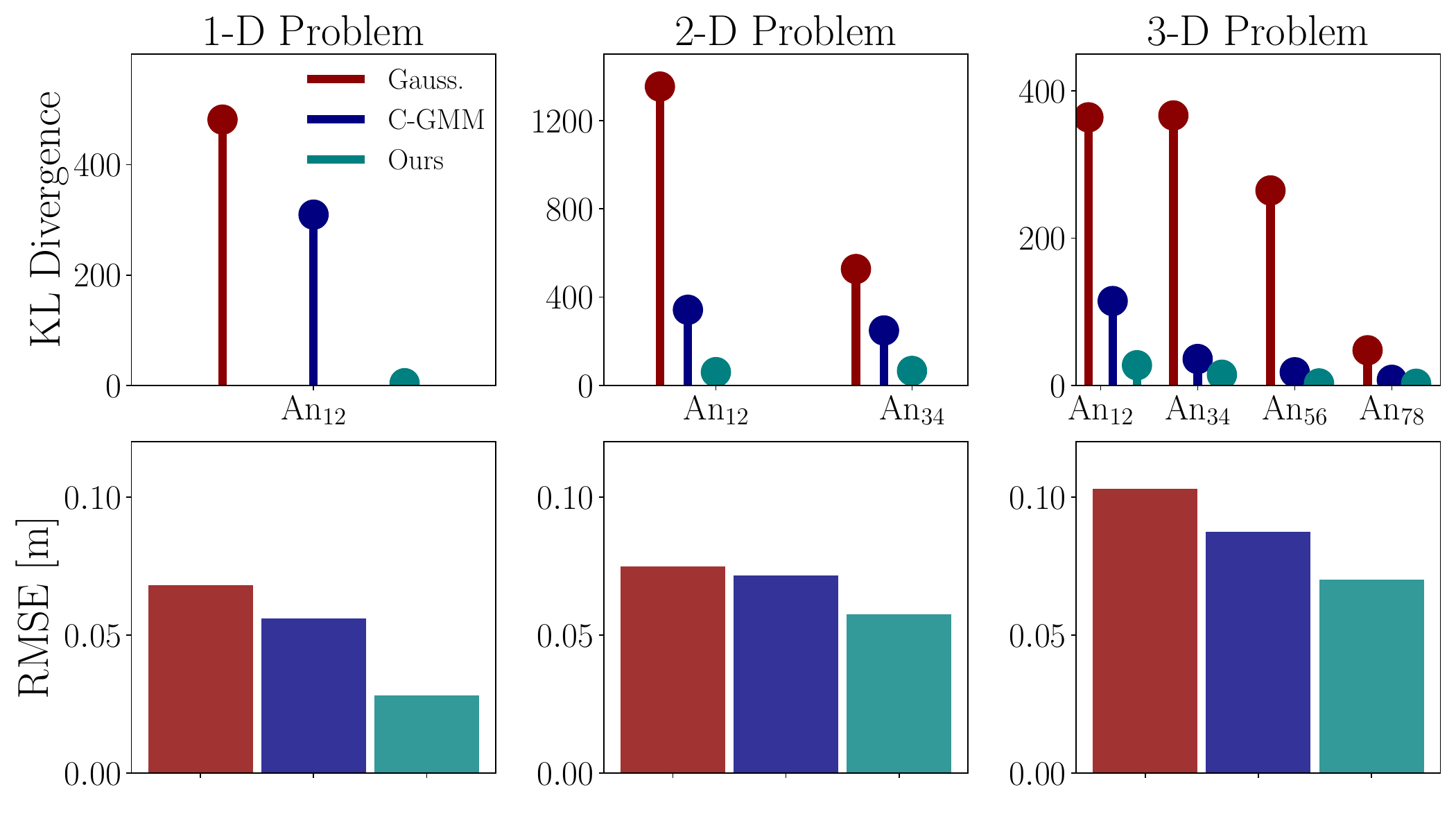}};
    \node[text width=1cm] at (-3.08, -2.35) {\scriptsize Gauss.};
    \node[text width=1cm] at (-2.37, -2.35) {\scriptsize C-GMM};
    \node[text width=1cm] at (-1.45, -2.35) {\scriptsize Ours};
    \node[text width=1cm] at (-0.3, -2.35) {\scriptsize Gauss.};
    \node[text width=1cm] at (0.4, -2.35) {\scriptsize C-GMM};
    \node[text width=1cm] at (1.35, -2.35) {\scriptsize Ours};
    \node[text width=1cm] at (2.55, -2.35) {\scriptsize Gauss.};
    \node[text width=1cm] at (3.25, -2.35) {\scriptsize C-GMM};
    \node[text width=1cm] at (4.2, -2.35) {\scriptsize Ours};
    \end{tikzpicture}
  \caption{KL-divergence values between the estimated noise model and the simulated noise model (top) and localization root-mean-square errors (RMSE) for 1-D, 2-D, and 3-D simulated problems. The proposed algorithm achieves better noise modeling performance and lower localization RMSE. }
  \label{fig:sim_res}
\end{figure}
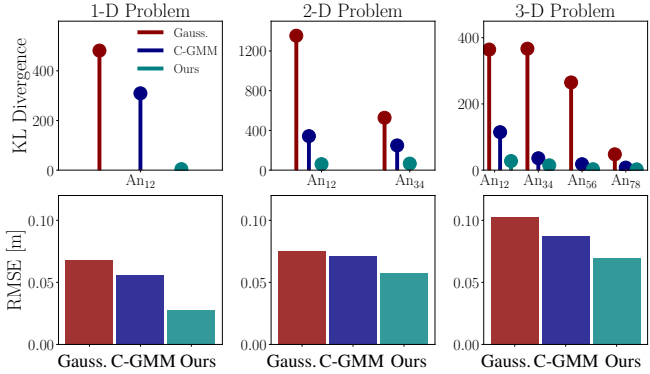

We initialize the proposed algorithm with odometry dead-reckoning and then apply the iterative bi-level optimization until convergence. We compare to two alternative noise modeling methods (1) Gaussian bias modeling, referred to as \textit{Gauss.}, which assumes the measurement residuals follow a single Gaussian distribution and (2) conventional GMM modeling, referred to as \textit{C-GMM}, that uses the measurement residuals directly for GMM parameter learning (similar to the methods presented in~\cite{pfeifer2019expectation,pfeifer2019incrementally}). Note we also use the terms \textit{Gauss.} and \textit{C-GMM} to represent the conventional joint localization and noise model learning methods using the corresponding noise modeling approaches. 
%
%
To verify the model learning performance, we compute the KL-divergence between the estimated noise models and the simulated ground truth GMM models. We demonstrate both the KL-divergence and the overall localization performance in Figure~\ref{fig:sim_res}. In the 1-D and 2-D problems, we can exactly propagate the uncertainty of the state through the linear motion models. Therefore, the proposed algorithm is initialized with the true state uncertainty information. We observe that our proposed noise modeling method (U-GMM) is able to achieve better GMM noise models (lower KL-divergence) and around $41.10\%$ and $34.78\%$ localization root-mean-squared error (RMSE) reductions compared to Gauss. and C-GMM, respectively.

\begin{figure}[b]
\centering
    \begin{tikzpicture}
    \definecolor{candyapplered}{rgb}{1.0, 0.03, 0.0}
    \node[inner sep=0pt, opacity=0.9] (anchor) at (0.0,0.0)
    {\includegraphics[width=.243\textwidth]{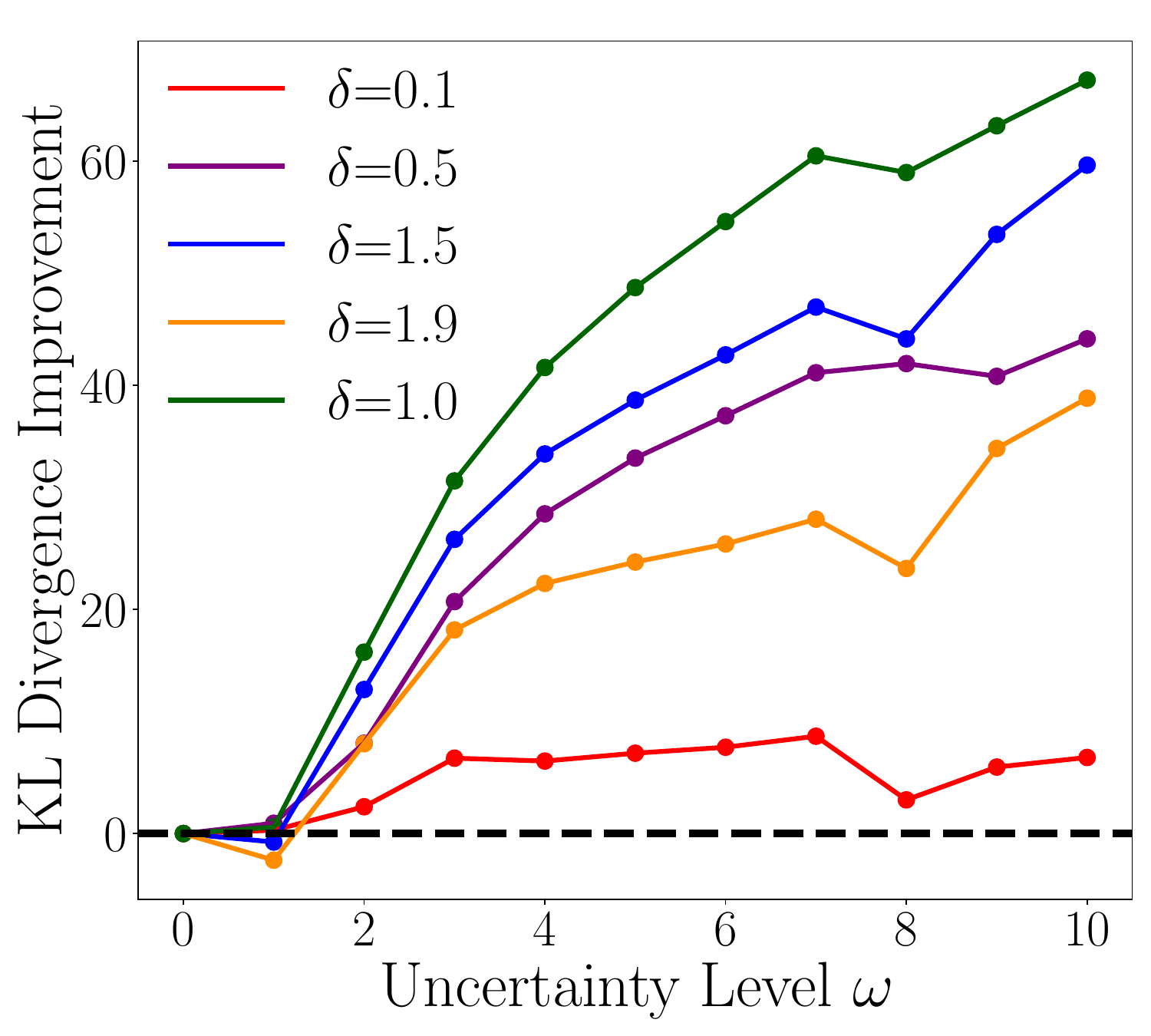}};
    \node[inner sep=0pt, opacity=0.9] (anchor) at (4.4,0.0)
    {\includegraphics[width=.246\textwidth]{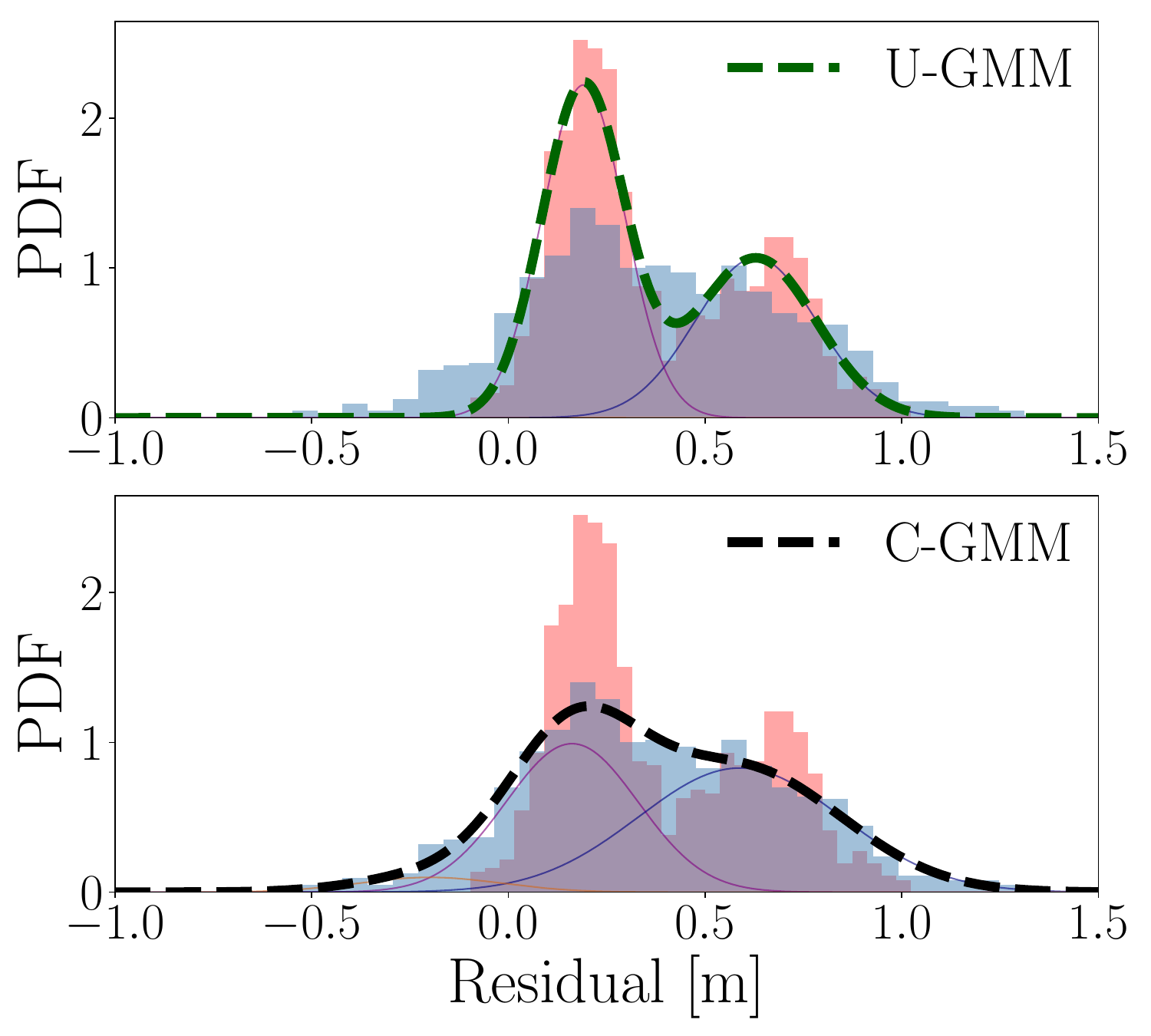}};
    \definecolor{amaranth}{rgb}{0.9, 0.17, 0.31}
    \definecolor{Blue}{rgb}{0.01, 0.28, 1.0}
    \draw[Blue, thick, dash pattern=on 1.5pt off 1.5pt] (5.1,0.7) -- (5.55,1.1);
    \node[text width=1cm, text=Blue] at (6.16, 1.3) {\scriptsize noisy};
    \node[text width=1cm, text=Blue] at (6.08, 1.1) {\scriptsize residual};
    \draw[amaranth, thick, dash pattern=on 1.5pt off 1.5pt] (4.0,1.55) -- (4.45, 1.4);
    \node[text width=3cm, text=amaranth] at (4.56, 1.73) {\scriptsize noise-free};
    \node[text width=3cm, text=amaranth] at (4.70, 1.5) {\scriptsize residual};
    \end{tikzpicture}
    \caption{The KL-divergence improvements of the proposed uncertainty-aware GMM (U-GMM) with uncertainty levels ranging from zero to ten compared to the conventional GMM (C-GMM), (left). The proposed U-GMM consistently outperforms C-GMM. The noise modeling performance comparison of U-GMM and C-GMM with $\omega=5$ is shown on the right.} 
    \label{fig:vi_model}
\end{figure}
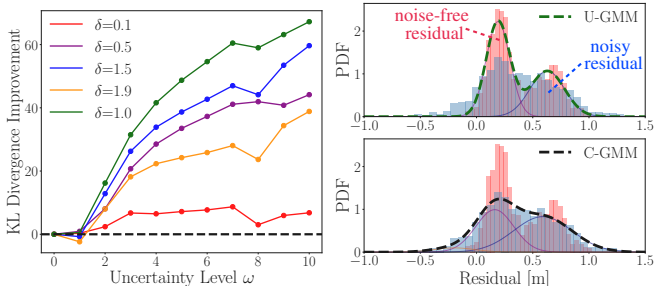
We then evaluate our algorithm on a simulated 3-D quadrotor pose estimation problem. The quadrotor is assumed to be equipped with a visual-inertial odometry (VIO) and an UWB tag. Four pairs of anchors $\Gamma=\{(1,2),(3,4),(5,6),(7,8)\}$ are simulated for TDOA localization. We use the standard \textit{special orthogonal group} \textit{SO}(3) to represent the rotation and \textit{special Euclidean group} \textit{SE}(3) to represent the robot's pose (rotation and translation). 
The robot pose $\mathbf{T}_{\mathcal{IB}}$ w.r.t. the inertial frame $\mathcal{F}_{\mathcal{I}}$ is parameterized as $\mathbf{T}_{\mathcal{IB}} = \{\mathbf{C}_{\mathcal{IB}}, \mathbf{p}_{\mathcal{I}}^{\mathcal{BI}}\} \in \textit{SE}(3)$ with the robot position $\mathbf{p}_{\mathcal{I}}^{\mathcal{BI}} \in \mathbb{R}^3$ and orientation $\mathbf{C}_{\mathcal{IB}} \in \textit{SO}(3)$. For brevity, we drop frame indicators in the following equations. We represent the uncertainties on 3-D poses with the right perturbation convention~\cite{sola2018micro}, leading to the following motion model: 
\begin{equation}
\small
    \mathbf{T}_{t} = \mathbf{T}_{t-1} \Delta\mathbf{T}_{t} \exp{(\bm{\xi}^{\wedge}_{t})}, 
\end{equation}
where the odometry noise is indicated as a pose perturbation $\bm{\xi}_t \sim \mathcal{N}(\bm{0},\bm{\Sigma}_u)$ and the $\wedge$ is the \textit{hat} operator that maps $\bm{\xi}_t\in\mathbb{R}^{6}$ to an element of the \textit{Lie algebra} $\mathfrak{se}(3)$. The corresponding UWB TDOA measurement model is 
\begin{equation}
\small
d_{ij,t_n} = \|\mathbf{C}_{t_n} \bm{l}_{ub} + \mathbf{p}_{t_n} -\bm{a}_j\| - \|\mathbf{C}_{t_n} \bm{l}_{ub} + \mathbf{p}_{t_n} -\bm{a}_i\| + \eta_{ij,t_n}, 
\end{equation}
where $\eta_{ij,t_n}$ is sampled from a simulated GMM distribution with two Gaussian components and $\bm{l}_{ub}$ is the position of the UWB radio in the body frame, also called \textit{lever arm}. We initialize the algorithm with the VIO dead-reckoning and follow the sigma-point transformation in~\cite{barfoot2014associating} for state uncertainty propagation. The KL-divergence values and the localization RMSE are shown in the third column of Figure~\ref{fig:sim_res}. Similarly, we observe that the proposed joint localization and uncertainty-aware noise model learning algorithm achieves better noise models and provides $31.98\%$ and $19.91\%$ localization RMSE reductions compared to Gauss. and C-GMM, respectively. 
%
\begin{table*}[htb]
\footnotesize
  \centering
  \setlength{\tabcolsep}{7.7pt}
  \renewcommand{\arraystretch}{1.8}
  \captionsetup{width=1.0\linewidth}
  \caption{Summary of the localization performance, shown by root-mean-square error (RMSE) in centimeters (cm), in the flying arena (Env. $\#1$). The trials are indicated as T for short. The visual-inertial odometry (VIO) dead-reckoning performance are shown in gray.}
  \begin{tabularx}{\linewidth}{>{\Centering}p{0.8cm} c c c c c c c c c c c c c c}
  \toprule
  \multirow{2}{*}{Alg.}     & \multicolumn{4}{c}{\makecell{Env. $\#1$, LOS \\ RMSE (cm)}}      &  \multicolumn{5}{c}{\makecell{Env. $\#1$, Cluttered Con. $\#1$\\ RMSE (cm)}}   & \multicolumn{5}{c}{\makecell{Env. $\#1$, Cluttered Con. $\#2$ \\ RMSE (cm)}}\\ 
       & T \#1  &T \#2   &T \#3    &T \#4   &T \#1    &T \#2   &T \#3  &T \#4  &T \#5   &T \#1    &T \#2   &T \#3  &T \#4  &T \#5  \\
  \midrule
   \rowcolor{myGray} 
   VIO    & 11.49    &13.94    &10.07     &15.39    &8.56     &12.26   &11.84       &14.09     &10.52     &8.51   &10.31   &7.38   &9.94     &12.88  \\
   Gauss. &6.45  &8.51  &8.65  &11.55  &6.22  &9.11  &9.26  &10.72  &8.23   &8.49   &9.33  &\textcolor{blue}{\textbf{5.75}}  &8.39  &11.77  \\
      &6.61  &7.22  &6.61  &9.78   &5.25  &6.33  &7.49  &9.92  &6.54  &5.87  &\textcolor{blue}{\textbf{7.84}}  &7.11 & \textcolor{blue}{\textbf{6.40}}   &7.86\\
   \textbf{Ours}  &\textcolor{blue}{\textbf{5.66}}   &\textcolor{blue}{\textbf{6.69}}  & \textcolor{blue}{\textbf{6.32}}   & \textcolor{blue}{\textbf{8.82}}   & \textcolor{blue}{\textbf{4.32}}  & \textcolor{blue}{\textbf{5.63}}  & \textcolor{blue}{\textbf{6.86}}  & \textcolor{blue}{\textbf{8.73}}  & \textcolor{blue}{\textbf{6.04}}  & \textcolor{blue}{\textbf{5.34}}  &8.07  & 6.94  & 6.99  & \textcolor{blue}{\textbf{7.33}}\\
  \bottomrule
  \end{tabularx}
    \begin{tikzpicture}[remember picture,overlay]
    \draw (-7.3, 3.0) -- (-3.0, 3.0);
    \draw (-2.6, 3.0) -- (2.9, 3.0);
    \draw (3.15, 3.0) -- (8.7, 3.0);
    \node[text width=2cm, text=black] at (-7.7, 0.9) {\footnotesize C-GMM};
    \end{tikzpicture}
  \label{tab:vicon_rmse}
\end{table*}

\subsection{U-GMM Noise Model Learning Evaluation}
\label{sec:model_verify}
We further evaluate the proposed U-GMM noise model learning method with different levels of state uncertainty. We demonstrate the simulation results on the 3-D problem setup while 1-D and 2-D results follow the same trend and can be found in the supplementary material\hyperlink{fn:supplementary}{$^1$}. We first simulated each estimated pose along the trajectory $\hat{\mathbf{T}}_n$ with pose perturbation $\bm{\xi}_n \sim \mathcal{N}(\bm{0}, \hat{\bm{\Sigma}}_{n})$, where $\hat{\bm{\Sigma}}_{n}=\textrm{diag}(\xi^{1}_{n}, \xi^2_{n}, \xi^{3}_{n}, \xi^{4}_{n}, \xi^5_{n}, \xi^{6}_{n})$ with $\{\xi^{1}_{n}, \xi^2_{n}, \xi^{3}_{n}\}$ and $\{\xi^{4}_{n}, \xi^5_{n}, \xi^{6}_{n}\}$ sampled from the uniform distribution $U(0, 0.035\omega)$ in meter and $U(0, 0.05\omega)$ in radian, respectively. The parameter $\omega \in\{0,\cdots, 10\}$ is the uncertainty level. We apply the proposed U-GMM method for noise modeling and visualize the KL-divergence improvements compared to C-GMM as the green curve in Figure~\ref{fig:vi_model} (left). One model performance comparison example with $\omega=5$ is shown in Figure~\ref{fig:vi_model} (right). We can observe that U-GMM consistently outperforms C-GMM in noise modeling performance. 
   
Moreover, as it is hard to access the exact uncertainty of the state in practice, we additionally evaluate the performance of U-GMM without precise state uncertainty information. We multiply the covariance matrix $\hat{\bm{\Sigma}}_{n}$ with $\delta \in \{0.1,0.5,1.5,1.9\}$ to simulate the scenarios when the state uncertainty is under- or over-estimated. We apply the same evaluation process and summarize the KL-divergence improvements in Figure~\ref{fig:vi_model} (left). It can be observed that even with inaccurate state uncertainty U-GMM outperforms the conventional method in almost all testing scenarios.

\subsection{Experimental Results for Indoor Localization}
We conducted extensive real-world experiments on a quadrotor platform for 3-D pose estimation to validate our proposed method. Two different cluttered environments, an indoor flying arena (Env. $\#1$) and a cafeteria (Env. $\#2$), were selected for the experiments. We used eight low-cost DW1000 UWB radios from Bitcraze as anchors to set up the UWB TDOA-based localization system. The UWB anchors were set to communicate in a round-robin network topology with $\Gamma=\{(8,1),(1,2),\cdots,(7,8)\}$. Our experimental platform is a customized quadrotor equipped with a DW1000 UWB radio and an Intel Realsense T261 stereo camera which provides odometry measurements. The T261 camera only runs visual-inertial odometry (VIO) as we disabled the map building and the loop closure feature during experiments. 

\begin{figure}[b]
\centering
    \begin{tikzpicture}
    \definecolor{candyapplered}{rgb}{1.0, 0.03, 0.0}
    \node[inner sep=0pt, opacity=0.9] (anchor) at (0.0,0.0)
    {\includegraphics[width=.245\textwidth]{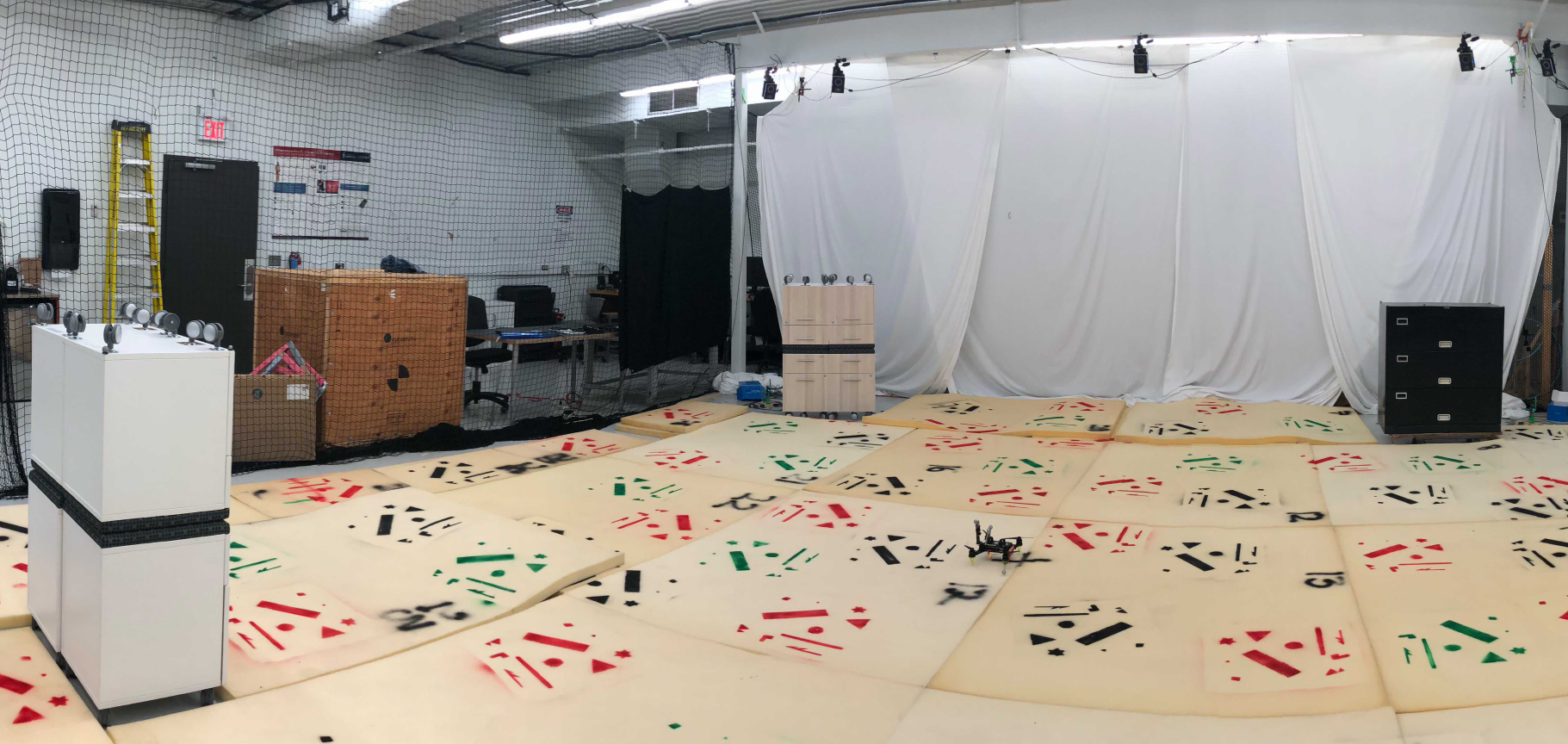}};
    \node[text width=4.5cm, text=black] at (0.2,-1.3) {(a) {\small Env. \#1, Cluttered Con. \#1}};
    \node[inner sep=0pt, opacity=0.9] (anchor) at (4.5,0.0)
    {\includegraphics[width=.245\textwidth]{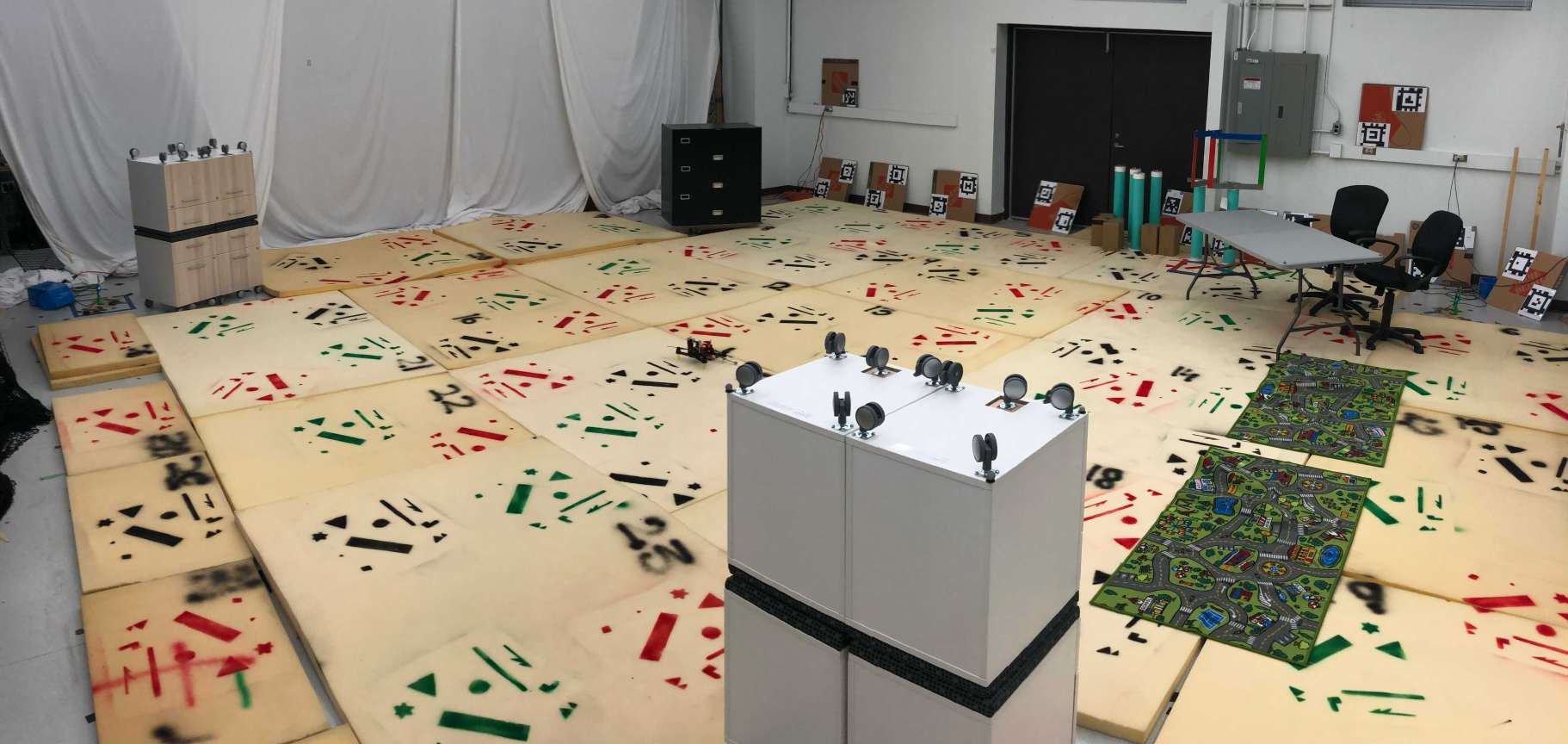}};
    \node[text width=4.5cm, text=black] at (4.7,-1.3) {(b) {\small Env. \#1, Cluttered Con. \#2}};
    \draw[candyapplered, dashed, very thick, rounded corners=1, shift={(5.25,0.0)},scale=0.90] (1.5,0) rectangle (0,1.05) node[right=-10pt,above right]{};
    \end{tikzpicture}
    \caption{The two cluttered conditions in the flying arena (Env. $\#1$). The extra two chairs and a desk are highlighted with a red dashed box in (b). The anchor positions are shown in Figure~\ref{fig:system-diagram}.}
    \label{fig:vicon_lab}
\end{figure}

We first conducted flight experiments in a $7m\times 8m\times 3 m$ indoor flying arena (Env. $\#1$). The VIO and raw UWB measurements were collected during flights. The raw VIO provides the dead-reckoning pose estimation w.r.t. the initial VIO frame. During the experiments, we aligned the VIO frame to the inertial frame at the starting point and converted the VIO measurements into incremental odometry between consecutive time steps: $\{\Delta \mathbf{T}_1, \cdots, \Delta \mathbf{T}_t\}$ with $\Delta \mathbf{T}_{t} = \mathbf{T}_{t-1}^{-1}\mathbf{T}_{t}$. The ground truth pose of the quadrotor was provided by a motion capture system comprising of ten Vicon cameras. The flight experiments were conducted in three different conditions: (1) LOS condition, (2) cluttered condition $\#1$ (with one metal and two wooden obstacles, see Figure~\ref{fig:vicon_lab}a), and (3) cluttered condition $\#2$ (with extra two chairs and one desk, see Figure~\ref{fig:vicon_lab}b). 
We initialized the algorithm with the VIO dead-reckoning and then performed the bi-level optimization until convergence for offline batch estimation. We conducted a total of $14$ trials of experiments in Env. $\#1$ and summarize the localization RMSE of the VIO dead-reckoning, Gauss., C-GMM, and our proposed algorithm in Table~\ref{tab:vicon_rmse}. In the four LOS experiments, the proposed method provides $20.04\%$ and $9.0\%$ RMSE reductions compared to Gauss. and C-GMM, respectively, leading to an average localization accuracy of $6.87$ centimeters. This performance improvement is due to the UWB systematic biases caused by hardware imperfection~\cite{zhao2021learning}. In Cluttered Con. $\#1$, our proposed algorithm demonstrates a similar localization accuracy ($6.31$ cm) as in LOS scenarios, resulting in $27.98\%$ and $11.39\%$ error reductions compared to Gauss. and C-GMM, respectively. However, in Cluttered Con. $\#2$, the proposed method achieves similar localization accuracy ($6.93$ cm) compared to C-GMM ($7.02$ cm). The reason is that the extra desk and chairs provided more visual features and helped to improve the VIO performance. This can be observed by that the average VIO dead-reckoning RMSE in Cluttered Con. $\#2$ is $9.8$ cm, which is lower than the other two conditions. As we demonstrated in Section~\ref{sec:model_verify}, our proposed method provides more improvements with larger state uncertainty. In contrast, when the state associated with small uncertainty due to less odometry drifts, both C-GMM and U-GMM result in similar localization performance. 
\begin{figure}[tb]
    \centering
    \begin{tikzpicture}
    \definecolor{springbud}{rgb}{0.65, 0.99, 0.0}
    \node[inner sep=0pt, opacity=0.9] (anchor) at (0.0,0.0)
    {\includegraphics[width=.45\textwidth]{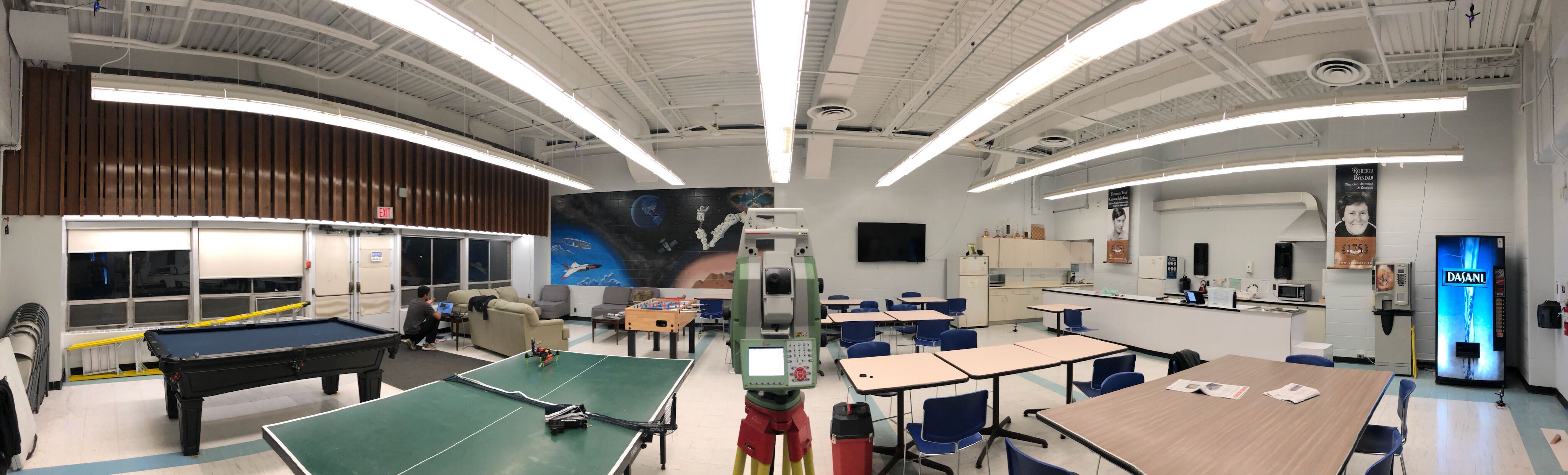}};
    \node[text width=3cm, text=blue] at (0.9,1.0) {\large{Env. \#2}};
    \node[circle, draw=blue, very thick, minimum size=5pt] (c) at (-1.05,0.5){};
    \node[circle, draw=blue, very thick, minimum size=5pt] (c) at (-3.5,1.2){};
    \node[circle, draw=blue, very thick, minimum size=5pt, dashed] (c) at (-1.1,-0.5){};
    \node[circle, draw=blue, very thick, minimum size=5pt, dashed] (c) at (-2.3,-0.99){};
    \node[circle, draw=blue, very thick, minimum size=5pt] (c) at (1.05,0.45){};
    \node[circle, draw=blue, very thick, minimum size=5pt] (c) at (1.17,-0.45){};
    \node[circle, draw=blue, very thick, minimum size=5pt] (c) at (3.65,-0.82){};
    \node[circle, draw=blue, very thick, minimum size=5pt] (c) at (3.5,1.15){};
    \node[inner sep=0pt, opacity=0.9] (anchor) at (0.0,-3.25)
    {\includegraphics[width=.45\textwidth]{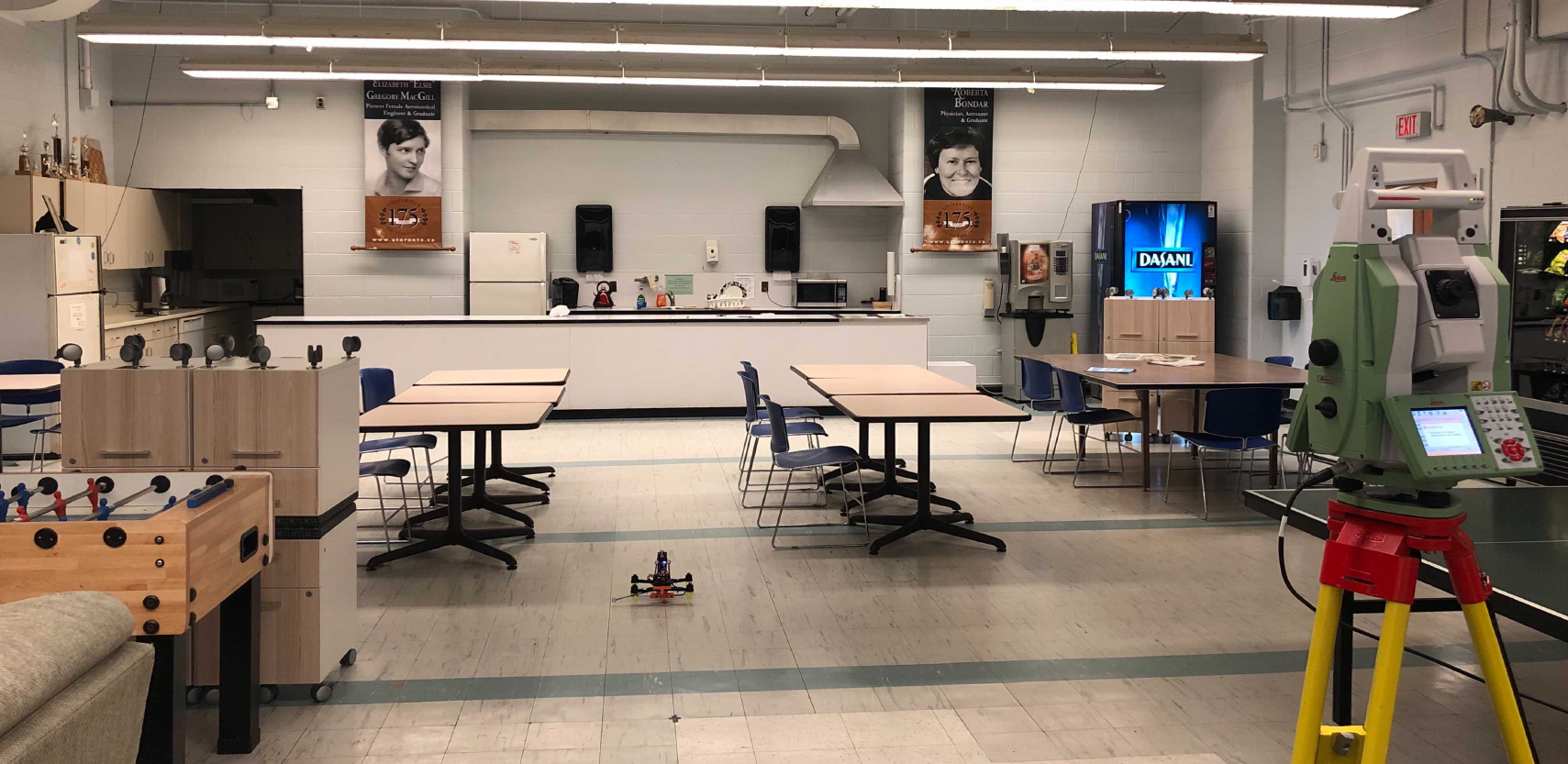}};
    \draw[red, ultra thick, rounded corners=1, shift={(1.45,-3.75)},scale=0.90] (1,0) rectangle (0,1.35) node[right=-10pt,above right]{};
    \draw[red, ultra thick, rounded corners=1, shift={(-3.8,-5.1)},scale=0.90] (2,0) rectangle (0,2.5) node[right=-10pt,above right]{};
    \end{tikzpicture}
    \caption{The cluttered cafeteria (Env. $\#2$) with the anchors and the extra wooden obstacles highlighted by blue circles and red boxes, respectively.}
    \label{fig:utias_cafe}
\end{figure}

\begin{figure}[t]
    \centering
    \includegraphics[width=.43\textwidth]{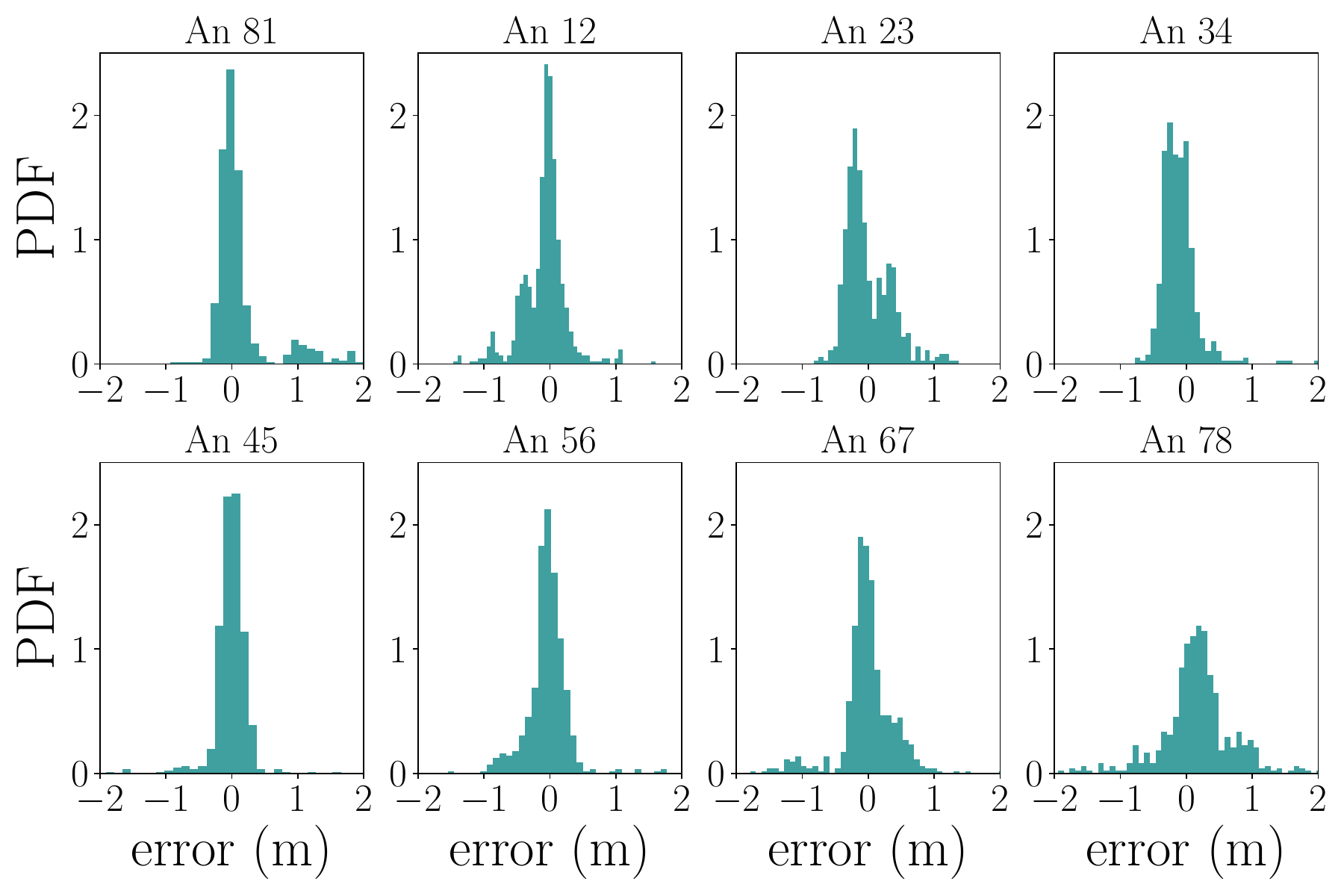}
  \caption{Histograms of UWB TDOA measurement residuals in the cluttered cafeteria. A significant amount of UWB anchors are occluded in this challenging environment and UWB residuals demonstrate non-Gaussian and multi-model distributions due to the NLOS and multi-path propagation.}
  \label{fig:error_hist}
\end{figure}

\definecolor{Gray}{rgb}{0.85,0.85,0.85}
\begin{table}[!b]
\footnotesize
  \centering
  \setlength{\tabcolsep}{6.75pt}
  \renewcommand{\arraystretch}{1.8}
  \captionsetup{width=1.0\linewidth}
  \caption{Localization performance shown by root-mean-square error (RMSE) in centimeters (cm), in a cluttered cafeteria. The trials are indicated as T for short. The visual-inertial odometry (VIO) dead-reckoning performance are shown in gray.}
  \begin{tabularx}{1.02\linewidth}{>{\Centering}p{0.5cm} c c c c c c c}
  \toprule
  \multirow{2}{*}{Alg.}     & \multicolumn{7}{c}{\makecell{Env. $\#2$, Cluttered Cafeteria, RMSE (cm)}}    \\
       & T \#1  &T \#2   &T \#3    &T \#4   &T \#5    &T \#6   &T \#7  \\
  \midrule
  \rowcolor{myGray} 
   \footnotesize{VIO}   & 41.66   & 30.13   & 26.19   & 25.92   & 24.66   & 16.63   & 34.19  \\
                  & 31.60    & 25.17   & 21.53   & 24.23   &19.65   & 23.02   & 29.61  \\
                  & 33.19   & 24.06   & 20.20   & 24.99   &\textcolor{blue}{\textbf{16.89}}  & 18.84   & 27.33 \\
   \textbf{\footnotesize{Ours}}  &\textcolor{blue}{\textbf{17.04}}   &\textcolor{blue}{\textbf{17.67}}   &\textcolor{blue}{\textbf{14.88}}   &\textcolor{blue}{\textbf{20.20}}  & 18.59   &\textcolor{blue}{\textbf{16.69}}   &\textcolor{blue}{\textbf{24.35}}  \\
  \bottomrule
  \end{tabularx}
    \begin{tikzpicture}[remember picture,overlay]
    \draw (-3.1, 3.0) -- (4.5, 3.0);
    \node[text width=2cm, text=black] at (-3.1, 1.5) {\footnotesize Gauss.};
    \node[text width=2cm, text=black] at (-3.25, 0.95) {\footnotesize C-GMM};
    \end{tikzpicture}
  \label{tab:cafe_rmse}
\end{table}
To further verify our proposed algorithm, we conducted seven trials of experiments in a $10m\times10m\times5m$ real-world cafeteria (Env. $\#2$, see Figure~\ref{fig:utias_cafe}). To create more NLOS conditions, we introduced two wooden obstacles (highlighted with red boxes in Figure~\ref{fig:utias_cafe}, bottom) to block the UWB anchors along with the default tables and chairs in the space. For ground truth data, we use a Leica total station in the tracking mode, which tracks the prism on the quadrotor and provides the position measurements at $5$ Hz. To quantify the localization accuracy, we manually moved the quadrotor at low speed to prevent the total station from losing track of the prism. The UWB measurement error histograms of one experiment are shown in Figure~\ref{fig:error_hist}. 
It can be observed that all eight UWB TDOA measurements were heavily corrupted due to this complex and cluttered environment. Similarly, we initialized the algorithm with the VIO dead-reckoning and performed the offline batch estimation. We summarize the localization performance of each experiment in the cafeteria in Table~\ref{tab:cafe_rmse}. The visual odometry performance degrades significantly in the cafeteria with an average dead-reckoning accuracy of $28.48$ cm. This is a very challenging environment as (1) the visual odometry drifts more due to the large space and (2) UWB measurements deteriorate greatly due to NLOS and multi-path radio propagation induced by different kinds of obstacles. In this cafeteria environment, our proposed method still achieves an average of $18.49$ cm localization accuracy, leading to $24.97\%$ and $19.11\%$ error reductions compared to Gauss. and C-GMM, respectively.

%% file: sections/6-conclusion.tex
\section{Conclusions}
In this work, we present a bi-level optimization-based localization and uncertainty-aware GMM noise model learning algorithm for UWB TDOA positioning systems. We explicitly incorporate the uncertainty of the estimated state into the GMM noise model learning to improve both noise modeling and localization performance. We demonstrate the effectiveness of our algorithm in numerous simulation scenarios and real-world experiments. In a laboratory setup, the proposed algorithm achieves an average of $6.70$ cm localization RMSE with low-cost UWB radios. We also evaluate our algorithm in a cluttered cafeteria and show that the proposed algorithm is able to achieve an average of $18.49$ cm localization accuracy, leading to $24.97\%$ and $19.11\%$ error reductions compared to conventional Gaussian and GMM-based methods, respectively.
